\documentclass[english]{ccdconf}
\usepackage[scaled=1.0]{helvet}
\usepackage{amsmath}
\usepackage{amsfonts}
\usepackage{graphicx}
\usepackage{cite}
\usepackage{bigstrut}

\begin{document}

\title{Sparse LiDAR Assisted Self-supervised Stereo Disparity Estimation}

\author{Xiaoming~Zhao\aref{irmct}, Weihai~Chen\aref{irmct}*, Xingming~Wu\aref{irmct}, Peter~C.~Y.~Chen\aref{nus}, and~Zhengguo~Li\aref{astar}*}


\affiliation[irmct]{School of Automation Science and Electrical Engineering, Beihang University, Beijing 100191, China \email{*whchenbuaa@126.com}}
\affiliation[nus]{Department of Mechanical Engineering, National University of Singapore, Singapore	\email{mpechenp@nus.edu.sg}}
\affiliation[astar]{SRO department, Institute for Infocomm Research, 1 Fusionopolis Way, Singapore \email{*ezgli@i2r.a-star.edu.sg}}

\maketitle
	
\begin{abstract}
Deep stereo matching has made significant progress in recent years. However, state-of-the-art methods are based on expensive 4D cost volume, which limits their use in real-world applications. To address this issue, 3D correlation maps and iterative disparity updates have been proposed. Regarding that in real-world platforms, such as self-driving cars and robots, the Lidar is usually installed. Thus we further introduce the sparse Lidar point into the iterative updates, which alleviates the burden of network updating the disparity from zero states. Furthermore, we propose training the network in a self-supervised way so that it can be trained on any captured data for better generalization ability. Experiments and comparisons show that the presented method is effective and achieves comparable results with related methods.
\end{abstract}


\footnotetext{This work was supported by the National Nature Science Foundation of China under Grant No. 61620106012, Beijing Natural Science Foundation under Grant No. 4202042, and the Key Research and Development Program of Zhejiang Province under Grant No. 2020C01109.}

\section{Introduction}

Estimating disparity/depth from stereo images is a common task in many applications. In recent years, the deep disparity estimation has get much progress, and yields better results than traditional handcrafted stereo algorithms. The state-of-the-art disparity estimation networks \cite{psmnet,cspn,adastereo} usually have four steps: 1) image feature extraction, 2) 4D cost volume generation, 3) cost aggregation, and 4) disparity regression. In this pipeline, the 4D cost volume is constructed by concatenating deep features on possible disparities, and computational expensive 3D convolutions are used to aggregate the 4D cost volume. Although simper aggregation operators are proposed \cite{ganet}, the high computational cost on 4D cost volume and the requirement for large labeled data still limit their use on practical platforms.

In this paper, we seek a solution to relieve the computational cost and labeled data requirements for stereo estimation networks. Given that Lidar is very likely to be installed on real-world platforms, collecting stereo images and Lidar data can be quick and easy. So we concentrate on self-supervised stereo disparity estimation with Lidar data.

A straight-forward way to improve stereo estimation with Lidar data is to fuse them into estimated stereo depth \cite{parkHighPrecisionDepthEstimation2020,zhaoLiDARToFBinocularDepthFusion2020}. However, the Lidar data cannot be used during stereo matching in this post-fusion strategy. To incorporate the Lidar data into the stereo matching, some researches \cite{poggiGuidedStereoMatching2019,wang3dLidarStereo2019} modify the cost volume based on Lidar data, while others \cite{listereo,choeVolumetricPropagationNetwork2021} fuse the Lidar feature into image feature. 
Nonetheless, they still rely on the expensive 4D cost volume. Recent works \cite{optstereo,raftstereo} inspired by the correlation and iterative pipeline in optical-flow estimation \cite{raft}, in which the 3D correlation map is constructed by correlating features on possible disparities, are very efficient and effective. In this paper, we propose incorporating Lidar points into iterative updates based on this pipeline. Specifically, the sparse disparity from the Lidar are regard as seminal states for the iteration. We adopt the convolutional spatial propagation network (CSPN) \cite{cspn} to estimate the pixel affinity, and propagate the sparse disparity with the affinity to a wider region after each iteration. This strategy allows the iterative steps to be reduced while improving the inference accuracy, thus reducing the running time.

\begin{figure*}[!tb]
	\includegraphics[width=\linewidth]{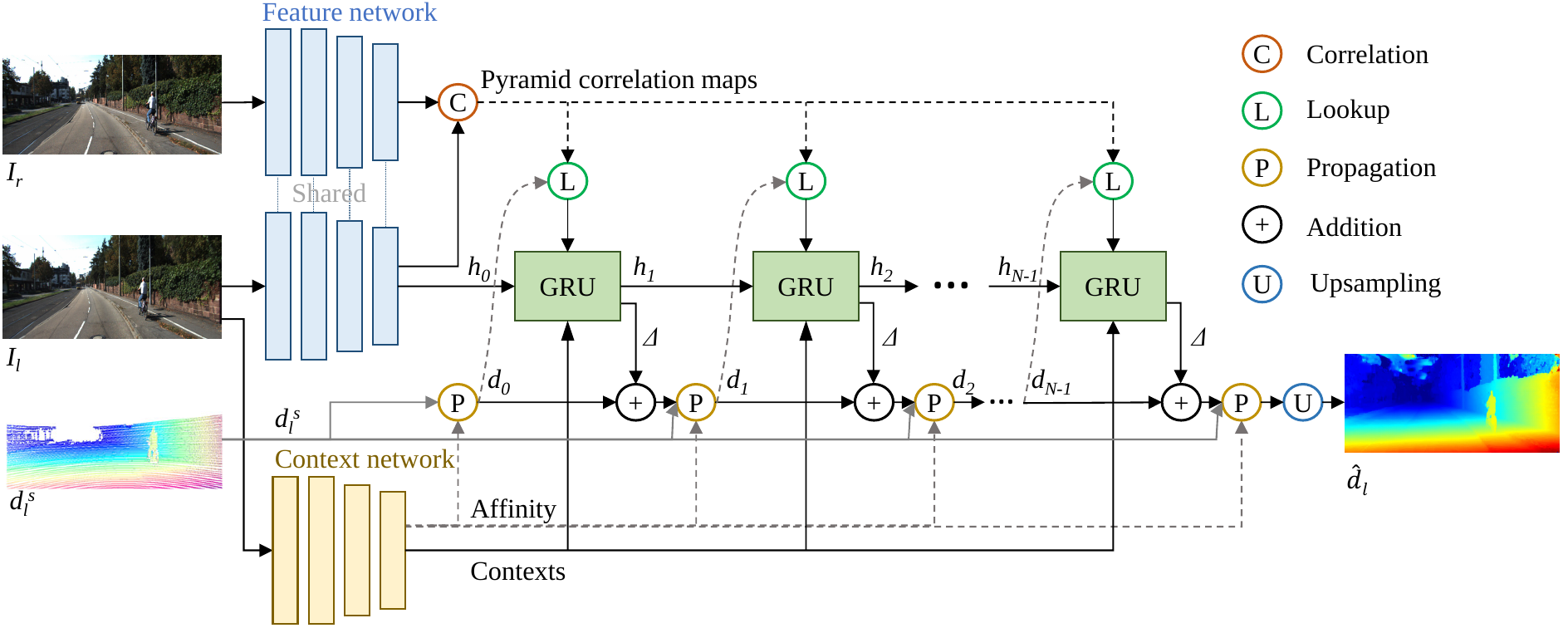}
	\caption{The proposed network takes a stereo image pair and left sparse disparity as input and outputs left dense disparity.}
	\label{fig:net}
\end{figure*}

Another limitation for supervised disparity estimation is the requirement of dense ground-truth disparity/depth label. Because acquiring dense depth data is costly and limited by the environment. The sun light would interfere with most dense depth sensors, thus we can only capture dense depth in the indoor environment. Therefore, only indoor or synthetic datasets can be used to train the network, which greatly limited the generalization ability. Fortunately, Lidar can operate without these constraints, thus, self-supervised training with stereo image and Lidar data is essential. We adopt advances of self-supervised optical flow estimation \cite{faminet, wangOcclusionAwareUnsupervised2018} to train the proposed network. 

To summarize, we develop a sparse Lidar-assisted stereo disparity estimation network by incorporating sparse depth into an iterative disparity update pipeline. As a result, the update steps can be reduced to improve running performance. In addition, self-supervised losses are used for network training. Therefore, the network can be trained on any stereo-Lidar data for improved generalization.

\section{Overview}
\label{method}
The goal of this paper is to estimate dense disparity with stereo images $\mathbf{I}_l$, $\mathbf{I}_r$ and the sparse Lidar depth $\mathbf{dep}_l^s$ of left image. For convenience, the sparse depth $\mathbf{dep}_l^s$ is first converted to sparse disparity $\mathbf{d}_l^s=fB/\mathbf{dep}_l^s$, where $f$ and $B$ denote the camera focal length and stereo baseline respectively. The proposed network takes $\mathbf{I}_l$, $\mathbf{I}_r$ and $\mathbf{d}_l^s$ as inputs and outputs the dense disparity of left image $\hat{\mathbf{d}}_l$. In the following, we will detail the network architecture and the training losses.
\section{Sparse assisted stereo estimation network}
The overall architecture of the propose network is shown in Fig. \ref{fig:net}, it mainly contains three stages: 1) the image feature and context extraction, 2) the correlation pyramid construction, and 3) the iterative disparity update.
\subsection{Image feature and context extraction}
The feature network extracts deep feature maps $\mathbf{f}_l,\ \mathbf{f}_r\in \mathbb{R}^{H\times W\times 128}$ for the left and right images $\mathbf{I}_l$, $\mathbf{I}_r$. For computational considerations, the feature maps are 1/4 or 1/8 of the original image resolution. They are then used to construct the correlation pyramid. The context network has the same architecture with the feature network, but with different network parameters. In contrast to RAFT-Stereo \cite{raftstereo}, we use an additional Residual block \cite{resnet} for image-dependent affinity estimation to optimize the estimated disparity with sparse input. The affinity $\mathbf{A}$ is a pixel by pixel kernels map for disparity propagation and optimization. In this paper, the kernel size is $3\times 3$, and the shape of this affinity map is $H\times W \times 8$.
\subsection{Correlation pyramid construction}
In cost volume based method, the 3D feature map $\mathbf{f}_l,\ \mathbf{f}_r\in \mathbb{R}^{H\times W\times C}$ are concatenated at different possible disparities. For example, when the maximum disparity is $D$, the shape of cost volume is $H\times W\times 2C \times D$, and the process of this 4D cost volume is computational expensive. On the other hand, the RAFT-Stereo \cite{raftstereo} shows that the 4D is not necessary, and the 3D correlation map is enough for disparity estimation. By computing the dot product of features at horizontal rows of stereo images (the ``C'' in Fig. \ref{fig:net}), the correlation map is constructed as
\begin{equation}
	\mathbf{C}(i,j,k)=\sum_{h} \mathbf{f}_l(i,j,h) \cdot \mathbf{f}_r(i,k,h), \quad \mathbf{C} \in \mathbb{R}^{H \times W \times W}.
\end{equation}
Following the RAFT-Stereo \cite{raftstereo}, the correlation pyramid is generated by 1D average pooling the last dimension of the correlation map with stride of 2. Thus the correlation map at level $k$ is of size $H\times W\times W/2^k$. In this pyramid, the top level provides a wide range of similarity, while the bottom level provides detailed local similarity.
\subsection{Iterative disparity update}
To update the disparity, the network first lookup (the ``L'' in Fig. \ref{fig:net}) into the correlation pyramid at current disparity estimation, and obtain the similarities at each pyramid level in a radius $r_L$ with bilinear sampling. Then the similarity and current disparity are encoded as motion features. With the context from context network, the delta disparity $\Delta_{k}$ is estimate from the multi-scale convolutional GRU after $k$-th iteration \cite{raftstereo}, and the disparity is updated as $\mathbf{d}_k = \mathbf{d}_{k-1} + \Delta_{k}$. To introduce the sparse disparity, we adopt the convolution spatial propagation (CSPN) \cite{cspn} before the first and after each GRU iteration 
\begin{equation} \small
\mathbf{d}_k = \operatorname{CSPN}(\mathbf{d}_{k-1} + \Delta_{k}\ |\ \mathbf{A}),
\end{equation}
where $\operatorname{CSPN}(\mathbf{d}\ |\ \mathbf{A})$ denotes the convolution spatial propagation on $\mathbf{d}\in \mathbb{R}^{H\times W}$ with the affinity map $\mathbf{A}\in \mathbb{R}^{H\times W \times 8}$ estimated from context network. On the affinity map, the last dimension denotes the eight neighborhood affinities of a location. Thus it can be reviewed as a $H\times W \times 3 \times 3$ tensor with the center pixel is zero. Then the disparity $\mathbf{d}$ is updated using the following equation
\begin{equation} \small
	\label{equ:aff}
	\begin{aligned}
		\mathbf{d}_{t+1}(i, j)=& \hat{\mathbf{A}}(i,j,0,0) \odot \mathbf{d}_{0}(i, j)+ \\ &\sum_{a, b=-1}^{a, b=1} \hat{\mathbf{A}}(i, j, a, b) \odot \mathbf{d}_{t}(i+a, j+b),
	\end{aligned}
\end{equation}
where
\begin{equation} \small
	\label{equ:aff1}
	\begin{aligned}
		& \hat{\mathbf{A}}(i, j, a, b)=\frac{\mathbf{A}_(i, j, a, b)}{\sum_{a, b=-1; a, b \neq 0}^{a,b=1}\left|\mathbf{A}(i, j, a, b)\right|} \\
		& \hat{\mathbf{A}}(i, j, 0,0)=\mathbf{1}-\sum_{a, b=-1; a, b \neq 0}^{a,b=1} \hat{\mathbf{A}}(i, j, a, b),
	\end{aligned}
\end{equation}
and $t$ is the update times of convolution spatial propagation, $0<i\le H$ and $0<j\le W$ denote the pixel location, $0<a\le 3$ and $0<b\le 3$ represent index of local $3\times 3$ kernel, and $\odot$ is an element-wise product. After each propagation, the sparse disparity is then applied to guarantee  that the disparities at valid positions $\mathbf{V}_l^s$ stay the same
\begin{equation} \small
	\mathbf{d}_{t+1}(i, j) = \left(1-\mathbf{V}_l^s(i,j)\mathbf{d}_{t+1}(i, j)\right) + \mathbf{V}_l^s(i,j)\mathbf{d}^s(i,j).
	\label{equ:spup}
\end{equation}

At last, the convex upsampling \cite{raftstereo} upsamples the low resolution disparity map to the original image resolution.

\section{Self-supervised loss functions}
Current stereo estimation networks frequently use different network parameters for different datasets, indicating a lack of generalizability. One approach to address this issue is to train on as many datasets as possible. The requirements for dense disparity ground-truth, on the other hand, make training on different datasets difficult. With the help of sparse Lidar input, we use self-supervised network training, which eliminates the need of dense disparity ground-truth.

\subsection{Appearance loss}
This loss is based on the image reconstruction difference. Considering the estimated dense disparity $\hat{\mathbf{d}}_l$ and $\hat{\mathbf{d}}_r$ for left and right image. The left image can be reconstructed with the right image $\mathbf{I}_r$
\begin{equation} \small
	\label{equ:warp}
	\hat{\mathbf{I}}_l\left(i,j\right) = \operatorname{Bisample}\left(\mathbf{I}_r\left(i,j-\hat{\mathbf{d}}_l\left(i,j\right)\right)\right),
\end{equation}
where $\operatorname{Bisample}\left(\mathbf{x}(i,j)\right)$ denotes the bilinear sampling at $(i,j)$ on image $\mathbf{x}$. The right image $\hat{\mathbf{I}}_r$ can also be reconstruct in the same way. Taking the left image as example, the appearance loss of left image \cite{monodepth}  is given as
\begin{equation}
	\small
	L_{ap}^{l} =  \frac{1}{N} \sum_{i, j \in 1-\mathbf{O}_l}\left( \alpha \frac{1-\operatorname{SSIM}\left(\mathbf{I}_{l}, \hat{\mathbf{I}}_{l}\right)}{2}+ (1-\alpha)\left\|\mathbf{I}_{l}-\hat{\mathbf{I}}_{l}\right\|\right),
	\label{equ:ap}
\end{equation}
where $\mathbf{O}_l$ is the occlusion map of left image based on range map \cite{wangOcclusionAwareUnsupervised2018}, $N$ denotes the number of non-occlusion pixels of left image, and $\operatorname{SSIM}\left(\mathbf{x},\mathbf{y}\right)$ computes the Structural Similarity (SSIM) \cite{ssim} of image $\mathbf{x}$ and $\mathbf{y}$, and $\alpha=0.85$ is a hyper-parameter to balance two loss terms \cite{monodepth}.

\subsection{Sparse disparity loss}
The sparse but accurate disparity can be used to not only help obtain the dense disparity, but also to supervise the disparity estimation. The sparse disparity supervision loss is deﬁned as follows:
\begin{equation} \small
	L_{sp}^{l} = \frac{1}{M}\sum_{i,j; \mathbf{V}_l^s(i,j)\ne 0} \left\| \mathbf{d}_l^s(i,j) - \hat{\mathbf{d}}_l(i,j) \right\|
\end{equation}
where $\mathbf{V}_l^s$ denotes the boolean valid map of sparse disparity $\mathbf{d}_l^s$, and $M$ denotes the number of valid points in $\mathbf{V}_l^s$.

\subsection{Left-Right consistency loss}
To guarantee that the estimated left disparity and right disparity are consistent, we apply the left-right disparity consistency loss. Taking the left disparity $\hat{\mathbf{d}}_l$ as example, right disparity $\hat{\mathbf{d}}_r$ is warped to left image with
\begin{equation} \small
	\hat{\mathbf{d}}_{r2l}=\operatorname{Bisample}\left(\hat{\mathbf{d}}_r\left(i,j-\hat{\mathbf{d}}_l\left(i,j\right)\right)\right).
\end{equation}
Then, the left-right consistency of left disparity is 
\begin{equation} \small
	L_{lr}^l = \frac{1}{N}\sum_{i, j \in 1-\mathbf{O}_l} \left\| \hat{\mathbf{d}}_{r2l}(i,j) - \hat{\mathbf{d}}_l(i,j) \right\|.
\end{equation}
where $\mathbf{O}_l$ and $N$ are defined in equation \eqref{equ:ap}. The consistency loss forces the corresponding positions between image pairs has the same disparity.

\subsection{Smooth loss}
Training with only appearance and sparsity supervise may result in noisy disparity estimation, which is non-smooth and inaccurate. By considering the structure of image, the smooth loss is introduced to alleviate this problem
\begin{equation} 
	\small
	L_{sm}^l=\frac{1}{N} \sum_{i, j \in 1-\mathbf{O}_l} \left(\left|\nabla_{x}^{2} \hat{\mathbf{d}}_l\right| e^{-\left|\nabla_{x}^{2} \mathbf{I}_{l}\right|}+
\left|\nabla_{y}^{2} \hat{\mathbf{d}}_l\right| e^{-\left|\nabla_{y}^{2} \mathbf{I}_{l}\right|}\right).
\end{equation}

\medspace

With all the losses presented, the total loss function for left image is given as
\begin{equation} \small
	L^l = \beta_{ap} L_{ap}^{l} + \beta_{sp} L_{sp}^{l} + \beta_{lr} L_{lr}^l+ \beta_{sm} L_{sm}^l,
\end{equation}
where $\beta$ represents the weight of each loss terms.
It is notable that predicting zero disparities is a trivial solution for left-right consistency loss and smooth loss. Therefore, their weight should not be too large, so as to avoid affecting appearance loss and sparse disparity loss, which play their main supervisory role.
In practice, the loss of right image is also considered, thus the overall loss is $L=(L^l+L^r)/2$.

\section{Experiments}
\label{exp}

\subsection{Datasets}
\textbf{SceneFlow} \cite{sceneflow} is a synthetic image dataset with 34,801 training frames that includes stereo image pairs, dense disparity, optical flow, . It contains three subsets: ``FlyingThings3D'', ``Monkaa', and ``Driving''. It also provides 4248 test frames on `FlyingThings3D' subset. This dataset is used for network ablation studies and network pre-training.

\textbf{KITTI} \cite{kitti} depth completion dataset provides stereo images, sparse Lidar depth, and semi-dense ground truth depth by accumulating successive 11 Lidar frames. The dataset consists of 42,949 training and 3,426 validation frames. This dataset also provides a selection of 1000 images for test, which are parts of the validation set. As the KITTI stereo benchmark dose not have Lidar data, and depth completion benchmark has no stereo image, so we did not submit the result. Following previous works \cite{listereo,choeVolumetricPropagationNetwork2021} the tests are conducted on the depth completion task.

\begin{figure}[!tb]
	\centering
	\includegraphics[width=0.9\linewidth]{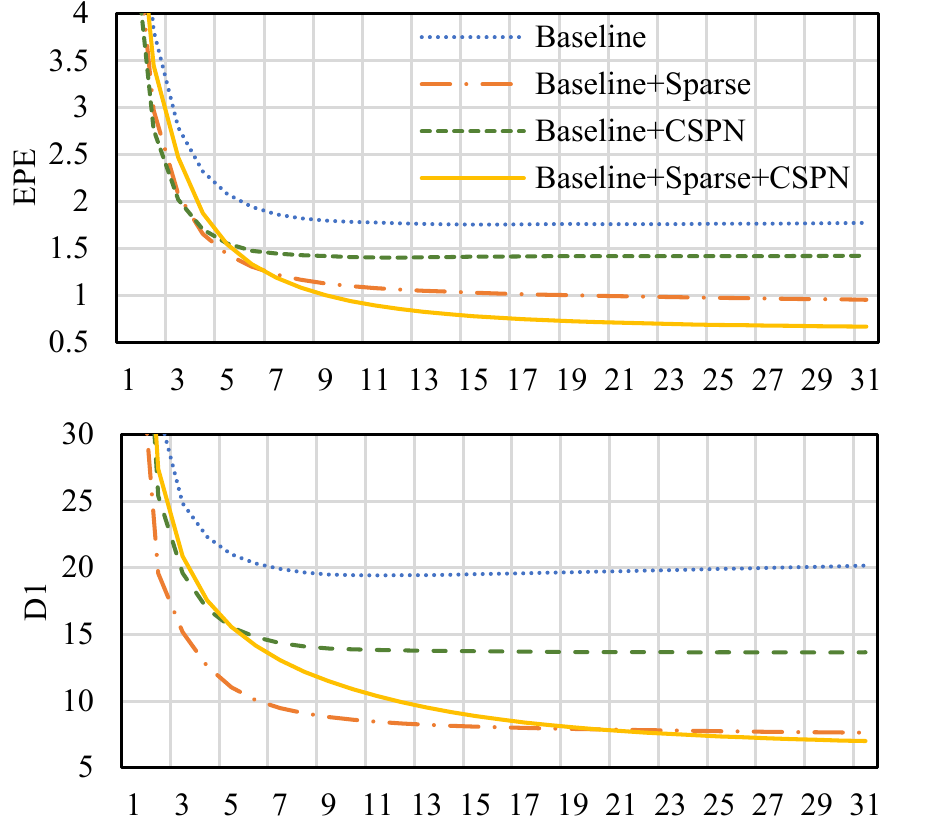}
	\caption{Ablation studies of network structure on FlyingThings3D test set \cite{sceneflow}. Top and bottom are the end-point-error (EPE) and D1 ($>1$px) with respect to GRU iteration numbers, respectively.}
	\label{fig:ablnet}
\end{figure}
\subsection{Implementation details}
\label{implementation}
The network is implemented with Pytorch and trained with one-cycle learning schedule \cite{smith2019super} with a maximum learning rate $2\times10^{-4}$. During the training, input images are cropped to $360\times720$, and are augmented with saturation augmentation (0~1.4). In the loss calculation, the $\beta_{ap}=1$, $\beta_{sp}=0.5$, $\beta_{lr}=0.01$, and $\beta_{sm}=0.01$. The details of ablation studies and comparisons are given below.

\textbf{Ablation studies} are conducted on the SceneFlow \cite{sceneflow} dataset to verify the proposed network architecture and self-supervise losses. In the training process, 500 random points are sampled from the ground truth depth to simulate the sparse Lidar data, the number of GRU and CSPN iterations are set to 10 times, and the batch size is 6. 

\textbf{The comparisons} with related works are conducted on the KITTI \cite{kitti} depth completion task. The training batch size is 8 in the comparisons, and both the GRU and CSPN iterations are 20 times. All LiDAR points are used as network inputs in supervised training, whereas in the self-supervised training, 250 points are network inputs and the remaining are used for sparse loss computation. During self-supervised training, the network is initialized with a self-supervised model trained on the SceneFlow \cite{sceneflow}. The RMSE (root mean square error), MAE (mean absolute error), iRMSE (inverse root mean square error in millimeter), and iMAE (inverse mean absolute error in millimeter) of estimated depth and ground-truth semi-dense depth are evaluated and compared.

\subsection{Ablation studies of network architecture}
We use the RAFT-Stereo \cite{raftstereo} as the baseline model. By introducing sparse disparity points and CSPN \cite{cspn} propagation module, four variations of networks are studied. To eliminate the effects of loss functions, we use supervised training of the network. We evaluate the end-point-error (EPE) of the estimated disparity and the percentage of disparity outliers of one pixel in first frame (D1) at 20K steps on the FlyingThings3D test set \cite{sceneflow}.
The evaluation results are shown in Fig. \ref{fig:ablnet}. The most obvious first phenomenon is that both the EPE and the D1 decrease as the number of iterations increases for all variations. However, the EPE and D1 of the baseline network can only be reduced to about 1.75 and 20\% respectively. By incorporating the CSPN \cite{cspn} propagation and sparse disparity, the EPE and D1 comes to 0.66 and 7\% respectively. This indicates that they are both helpful for disparity estimation. Morever, one can also use much lesser iterations to obtain equivalent disparity accuracy with the CSPN \cite{cspn} propagation and sparse disparity, which is important for efficient applications. Without the LiDAR measurements, extracting accurate sparse keypoints \cite{zhao2021alike} and matching them \cite{zhao2021probabilistic} would also be possible to act as the accurate sparse input.

\subsection{Ablation studies of self-supervise training}
\label{sec:ablloss}
\begin{table}[t]
	\centering
	\caption{Ablation studies of loss functions on FlyingThings3D test set \cite{sceneflow}. The ``App'' and ``Sps'' denote appearance and sparse loss respectively. The ``+CSPN'' and ``+Sparse'' represent network variations with CSPN and sparse respectively. And the ``-half1'' and ``-half2'' are detailed in Section \ref{sec:ablloss}.}
	\setlength{\tabcolsep}{0.5mm}
	{
		\begin{tabular*}{\linewidth}{@{}@{\extracolsep{\fill}}lccrr@{}}
			\hline
			\textbf{Network}                         & \textbf{App} & \textbf{Sps} &  \textbf{EPE} & \textbf{D1($>$1px)} \bigstrut \\ \hline
			\textbf{Baseline}                        &  \checkmark  &              &         11.48 &          26.51 \bigstrut[t] \\
			\textbf{Baseline}                        &  \checkmark  &  \checkmark  &          6.05 &          26.18 \bigstrut[b] \\ \hline
			\textbf{Baseline+CSPN}                   &  \checkmark  &              &         10.06 &          25.58 \bigstrut[t] \\
			\textbf{Baseline+CSPN+Sparse}            &  \checkmark  &  \checkmark  &          6.06 &                       25.87 \\
			\textbf{Baseline+CSPN+Sparse-half1}      &  \checkmark  &  \checkmark  &          1.57 &                       13.56 \\
			\textbf{Baseline+CSPN+Sparse-half2}      &  \checkmark  &  \checkmark  & \textbf{1.24} & \textbf{12.15} \bigstrut[b] \\ \hline
			\textbf{Baseline+CSPN+Sparse-Supervise} &      -       &      -       & \textbf{0.66} &  \textbf{7.00} \bigstrut[t] \\ \hline
		\end{tabular*}%
	}
	\label{tab:ablloss}%
\end{table}%

\begin{figure}[t]
	\centering
	\includegraphics[width=\linewidth]{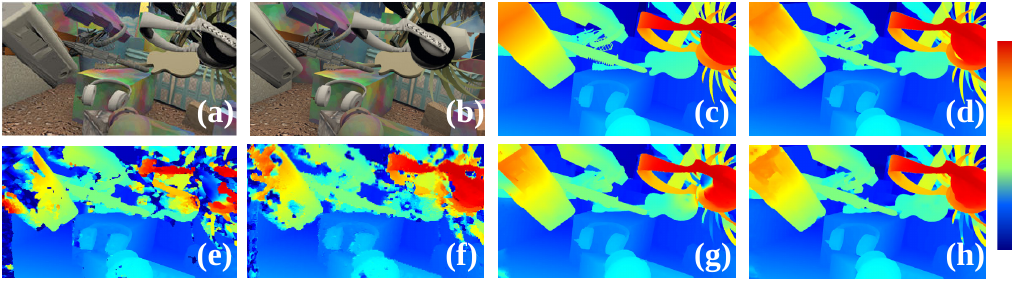}
	\caption{Visualization of evaluation results on FlyingThings3D \cite{sceneflow} test set. (a) left image, (b) right image, (c) left ground-truth disparity, (d)-(h) estimated left disparity of supervised baseline, self-supervised baseline without sparse,  self-supervised baseline+CSPN with sparse, self-supervised baseline+CSPN with sparse-half1, and self-supervised baseline+CSPN with sparse-half2, respectively.}
	\label{fig:ablloss}
\end{figure}

We investigated appearance and sparsity disparity loss on different network variations while keeping the weights of left-right consistency and smooth loss constant at 0.01. The test results are shown in Table \ref{tab:ablloss} and Fig. \ref{fig:ablloss}. First, the baseline network is tested for appearance and sparsity disparity loss. As shown in Table \ref{tab:ablloss}, adding sparse disparity improves the EPE (from 11.48 to 6.05) but has limited effect on the D1 (26.51\% and 26.18\%). When the sparse disparity is added to the baseline+CSPN variation, the EPE falls from 10.06 to 6.06, but the D1 remains high above 25\%. The visualization of estimated disparity in Fig. \ref{fig:ablloss} (e) and (f) also shows that the network has difficulty fitting large disparities.
After careful examination, we discover that the disparity update in Equation \eqref{equ:spup} produces the same disparity and ground-truth disparity of the corresponding sparse positions in the output dense disparity map (equivalent to directly replacing the disparity of effective sparse points with ground-truth disparity). As a result, the sparsity loss is masked out during network back propagation. To avoid this problem, we tested two self-supervised strategies with the sparse disparities: 
\begin{itemize}
\item[$\bullet$]\textbf{Sparse-half1}: Half of the sparse disparities is randomly selected as the network input, and all the sparse disparities is used to calculate the sparse loss; 
\item[$\bullet$]\textbf{Sparse-half2}: Half of the sparse disparities is randomly selected as the network input, and the remaining sparse disparities is used to calculate the sparse loss.
\end{itemize}

The EPE and D1 are reduced to 1.57 and 13.56\%, respectively, with the Sparse-half1 training strategy, as shown in Table \ref{tab:ablloss}. But it is inferior to the Sparse-half2 training strategy because half of the sparse points are still invalid in network back propagation and would affect affinity training. The Sparse-half2 training strategy reduces the EPE and D1 to 1.24 and 12.15\%, respectively, which is much closer to supervised training results. In addition, the visualization in Fig. \ref{fig:ablloss} confirms the efficacy of the proposed strategy.

\begin{table}[t]
	\centering
	\caption{Comparisons with related supervised methods on KITTI depth completion validation set. ``M+L'' and ``S+L'' denote ``Mono+LiDAR'' and ``Stereo+LiDAR'' respectively. The best and second best of ``S+L'' are \textbf{\textit{ITALIC}} and \textbf{BOLD}.}
	\setlength{\tabcolsep}{0.5mm}
	{
	\begin{tabular*}{\linewidth}{@{}@{\extracolsep{\fill}}clrrrr@{}}
		\hline
		\textbf{Modality} & \textbf{Method}                                         &            \textbf{RMSE} &             \textbf{MAE} &         \textbf{iRMSE} &    \textbf{iMAE} \bigstrut \\ \hline
		\textbf{M+L}      & \textbf{Sparsetodense \cite{sparsetodense}}                    &                   814.70 &                   249.90 &                   2.80 &          1.21 \bigstrut[t] \\
		\textbf{M+L}      & \textbf{RGB\_certainty \cite{van2019sparse}}                         &                   772.80 &                   215.00 &                   2.19 &                       0.93 \\
		\textbf{M+L}      & \textbf{Spade \cite{jaritz2018sparse}}                           &                  1035.29 &                   248.32 &                   2.60 &                       0.98 \\
		\textbf{M+L}      & \textbf{CSPN \cite{cspn}}                                        &                  1019.64 &                   279.46 &                   2.93 &                       1.15 \\
		\textbf{M+L}      & \textbf{Guidenet \cite{tang2020learning}}                        &                   777.78 &                   221.59 &                   2.39 &                       1.00 \\
		\textbf{M+L}      & \textbf{NLSPN \cite{park2020non}}                                &                   771.80 &                   197.30 &                   2.00 &                       0.80 \\
		\textbf{M+L}      & \textbf{PENet \cite{penet}}                                      &                   757.20 &                   209.00 &                   2.22 &                       0.92 \\
		\textbf{M+L}      & \textbf{ACMNet \cite{zhao2021adaptive}}                          &                   790.75 &                   217.34 &                   2.39 &          0.97 \bigstrut[b] \\ \hline
		\textbf{S+L}      & \textbf{Park et al. \cite{parkHighPrecisionDepthEstimation2020}} &                  2021.20 &                   500.50 &                   3.39 &          1.38 \bigstrut[t] \\
		\textbf{S+L}      & \textbf{CCVN \cite{wang20193d}}                                  & \textit{\textbf{749.30}} &          \textbf{252.50} & \textit{\textbf{1.40}} &     \textit{\textbf{0.81}} \\
		\textbf{S+L}      & \textbf{LiStereo \cite{listereo}}                                &                   832.16 &                   283.91 &          \textbf{2.19} &                       1.10 \\
		\textbf{S+L}      & \textbf{Ours}                                                    &          \textbf{775.66} & \textit{\textbf{210.26}} &                   2.23 & \textbf{0.86} \bigstrut[b] \\ \hline
	\end{tabular*}%
	}
	\label{tab:sup}%
\end{table}%

\begin{table}[t]
	\centering
	\caption{Comparisons with related self-supervised methods on KITTI depth completion validation set. ``M+L'' and ``S+L'' denote ``Mono+LiDAR'' and ``Stereo+LiDAR'' respectively. The best and second best are \textbf{\textit{ITALIC}} and \textbf{BOLD}.}
	\setlength{\tabcolsep}{0.5mm}
	{
	\begin{tabular*}{\linewidth}{@{}@{\extracolsep{\fill}}clrrrr@{}}
		\hline
		\textbf{Modality} & \textbf{Method}                      &             \textbf{RMSE} &             \textbf{MAE} &         \textbf{iRMSE} &    \textbf{iMAE} \bigstrut \\ \hline
		\textbf{M+L}      & \textbf{Sparsetodense \cite{sparsetodense}} &                   1301.05 &                   352.22 &          \textbf{4.08} &          1.61 \bigstrut[t] \\ \hline
		\textbf{S+L}      & \textbf{LiStereo \cite{listereo}}             & \textit{\textbf{1278.87}} & \textit{\textbf{326.10}} & \textit{\textbf{3.83}} &     \textit{\textbf{1.32}} \bigstrut[t] \\
		\textbf{S+L}      & \textbf{Ours}                                 &          \textbf{1293.56} &          \textbf{342.66} &                   4.48 & \textbf{1.34} \bigstrut[b] \\ \hline
	\end{tabular*}%
	}
	\label{tab:self}%
\end{table}%
\begin{figure}[t]
	\centering
	\includegraphics[width=\linewidth]{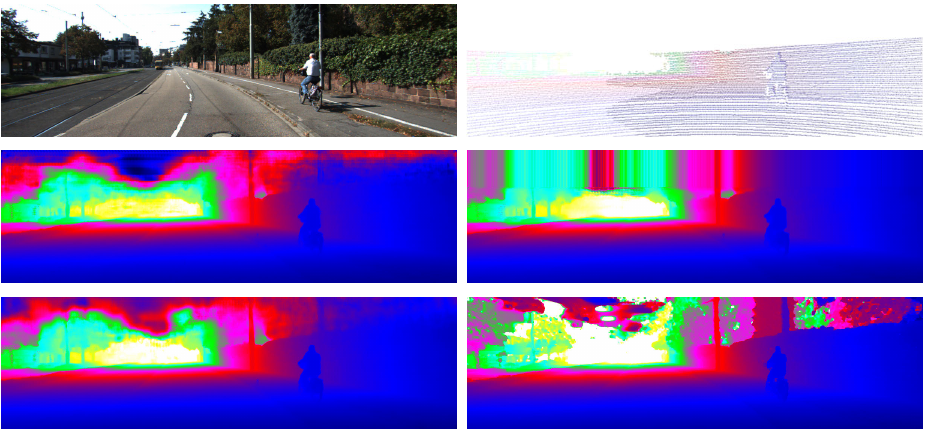}
	\caption{Visualization of evaluation results on KITTI depth completion \cite{kitti} selected validation set. The top two images are the left image and sparse LiDAR depth. And the below four images visualizes the depth of Sparse-to-dense \cite{sparsetodense}, RGB\_guide \cite{van2019sparse}, PENet \cite{penet} and our method.}
	\label{fig:kitti}
\end{figure}

\subsection{Comparison to supervised methods}
Table \ref{tab:sup} shows the comparisons with supervised methods. It can be seen that the proposed method achieves close to the best errors in the S+L methods, ranking first in MAE and second in RMSE and iMAE. However, we can see that some M+L methods have lower errors than S+L methods. This is due to the S+L methods rely heavily on accurate LiDAR points while images are primarily used for guidance, whereas M+L methods rely mainly on stereo matching of binocular images and sparse LiDAR points are only used for guidance or optimization. One disadvantage of S+L methods is that they cannot infer depth in areas with no LiDAR points, as shown in Fig. \ref{fig:kitti}.

\subsection{Comparison to self-supervised methods}
The comparisons with self-supervised methods are shown in Table \ref{tab:self}. We can see that the proposed method ranks second in terms of RMSE, MAE, and iMAE, and is very close to LiStereo \cite{listereo}, proving its effectiveness. More importantly, the proposed method is scalable: the network can be configured with or without sparse LiDAR input, as well as with a different number of GRU iterations for different computational budgets (Fig. \ref{fig:ablnet}). And it is also possible to feed the network stereo images with different exposures to generate high dynamic range 3D images \cite{kou2017intelligent}.

\section{Conclusions}
\label{conclusion}

We present a self-supervised stereo disparity estimation network with sparse disparity input in this paper. To incorporate the sparse disparity input to the stereo matching pipeline, the affinity of disparity is estimated. In each GRU iteration, convolutional spatial propagation with affinity and sparse disparity is used to optimize the disparity. The appearance difference and disparity on sparse points are also used to train the network in a self-supervised manner. We can train the network in any dataset using the self-supervised learning method, as there are no ground-truth label requirements. Our future work will include improving affinity propagation and introducing stereo images with different exposures for the generation of high dynamic range 3D images.


\begin{thebibliography}{10}	
	
	\bibitem{psmnet}
	J.-R. Chang and Y.-S. Chen, Pyramid stereo matching network, in
	Proceedings of the {IEEE} {Conference} on {Computer} {Vision} and
		{Pattern} {Recognition}, 5410--5418, 2018
	
	\bibitem{cspn}
	X.~Cheng, P.~Wang, and R.~Yang, Learning {Depth} with
		{Convolutional} {Spatial} {Propagation} {Network},
	IEEE transactions on pattern analysis and machine intelligence, Vol.~42, No.~10, 2361--2397, 2019.
	
	\bibitem{adastereo}
	X.~Song, G.~Yang, X.~Zhu, H.~Zhou, Z.~Wang, and J.~Shi, {AdaStereo}: a simple
	and efficient approach for adaptive stereo matching, in Proceedings
		of the {IEEE}/{CVF} {Conference} on {Computer} {Vision} and {Pattern}
		{Recognition}, 10328--10337, 2021
	
	\bibitem{ganet}
	F.~Zhang, V.~Prisacariu, R.~Yang, and P.~H. Torr, Ga-net: {Guided}
	aggregation net for end-to-end stereo matching, in Proceedings of the
		{IEEE}/{CVF} {Conference} on {Computer} {Vision} and {Pattern}
		{Recognition}, 185--194, 2019
	
	\bibitem{parkHighPrecisionDepthEstimation2020}
	K.~Park, S.~Kim, and K.~Sohn, High-{Precision} {Depth} {Estimation} {Using}
	{Uncalibrated} {LiDAR} and {Stereo} {Fusion}, IEEE Trans. on
		Intelligent Transportation Systems, Vol.~21, No.~1, 321--335, 2020.
	
	\bibitem{zhaoLiDARToFBinocularDepthFusion2020}
	X.~Zhao, W.~Chen, Z.~Liu, X.~Ma, L.~Kong, X.~Wu, H.~Yue, and X.~Yan,
	{LiDAR}-{ToF}-{Binocular} depth fusion using
		gradient priors, in 2020 {Chinese} {Control}
			{And} {Decision} {Conference}.
	Hefei, China, 2024--2029, 2020
	
	\bibitem{poggiGuidedStereoMatching2019}
	M.~Poggi, D.~Pallotti, F.~Tosi, and S.~Mattoccia,
	Guided {Stereo} {Matching}, in
	2019 {IEEE}/{CVF} {Conference} on {Computer}
			{Vision} and {Pattern} {Recognition}. Long Beach, CA, USA, 979--988, 2019
	
	\bibitem{wang3dLidarStereo2019}
	T.-H. Wang, H.-N. Hu, C.~H. Lin, Y.-H. Tsai, W.-C. Chiu, and M.~Sun, 3d lidar
	and stereo fusion using stereo matching network with conditional cost volume
	normalization, in {IEEE}/{RSJ} {International} {Conference} on
		{Intelligent} {Robots} and {Systems}, 5895--5902, 2019
	
	\bibitem{listereo}
	J.~Zhang, M.~S. Ramanagopal, R.~Vasudevan, and M.~Johnson-Roberson, Listereo:
	{Generate} dense depth maps from lidar and stereo imagery, in 2020
		{IEEE} {International} {Conference} on {Robotics} and {Automation}, 7829--7836, 2020
	
	\bibitem{choeVolumetricPropagationNetwork2021}
	J.~Choe, K.~Joo, T.~Imtiaz, and I.~S. Kweon, Volumetric propagation network:
	{Stereo}-lidar fusion for long-range depth estimation, IEEE Robotics
		and Automation Letters, Vol.~6, No.~3, 4672--4679, 2021.

	
	\bibitem{optstereo}
	H.~Wang, R.~Fan, P.~Cai, and M.~Liu, {PVStereo}: {Pyramid} {Voting} {Module}
	for {End}-to-{End} {Self}-{Supervised} {Stereo} {Matching}, IEEE
		Robotics and Automation Letters, Vol.~6, No.~3, 4353--4360, 2021.
	
	\bibitem{raftstereo}
	L.~Lipson, Z.~Teed, and J.~Deng, {RAFT}-{Stereo}: {Multilevel} {Recurrent}
	{Field} {Transforms} for {Stereo} {Matching}, arXiv preprint
		arXiv:2109.07547, 2021.
	
	\bibitem{raft}
	Z.~Teed and J.~Deng, {RAFT}: {Recurrent} {All}-{Pairs} {Field} {Transforms}
	for {Optical} {Flow}, in European {Conference} on {Computer}
		{Vision}, 2020.
	
	\bibitem{faminet}
	Z.~Liu, J.~Liu, W.~Chen, X.~Wu, and Z.~Li, Faminet: Learning real-time
	semi-supervised video object segmentation with steepest optimized optical
	flow, IEEE Trans. on Instrumentation and Measurement, Vol.~71,
	1--16, 2022.
	
	\bibitem{wangOcclusionAwareUnsupervised2018}
	Y.~Wang, Y.~Yang, Z.~Yang, L.~Zhao, P.~Wang, and W.~Xu, Occlusion aware
	unsupervised learning of optical flow, in Proceedings of the {IEEE}
		{Conference} on {Computer} {Vision} and {Pattern} {Recognition}, 4884--4893, 2018
	
	\bibitem{resnet}
	K.~He, X.~Zhang, S.~Ren, and J.~Sun, Deep {Residual}
		{Learning} for {Image} {Recognition}, in
	{IEEE} {Conference} on {Computer} {Vision}
			and {Pattern} {Recognition}. Las
	Vegas, USA, 770--778, 2016
	
	\bibitem{monodepth}
	C.~Godard, O.~Mac~Aodha, and G.~J. Brostow, Unsupervised monocular depth
	estimation with left-right consistency, in Proceedings of the {IEEE}
		conference on computer vision and pattern recognition, 270--279, 2017
	
	\bibitem{ssim}
	Z.~Wang, A.~C. Bovik, H.~R. Sheikh, and E.~P. Simoncelli, Image quality
	assessment: from error visibility to structural similarity, IEEE
		transactions on image processing, Vol.~13, No.~4, 600--612, 2004.
	
	\bibitem{sceneflow}
	N.~Mayer, E.~Ilg, P.~Hausser, P.~Fischer, D.~Cremers, A.~Dosovitskiy, et.al, A large dataset to train convolutional networks for disparity,
	optical flow, and scene flow estimation, in Proceedings of the {IEEE}
		conference on computer vision and pattern recognition, 4040--4048, 2016
	
	\bibitem{kitti}
	A.~Geiger, P.~Lenz, C.~Stiller, and R.~Urtasun, Vision meets robotics: The
	kitti dataset, International Journal of Robotics Research, 2013.
	
	\bibitem{smith2019super}
	L.~N. Smith and N.~Topin, Super-convergence: Very fast training of neural
	networks using large learning rates, in Artificial Intelligence and
		Machine Learning for Multi-Domain Operations Applications, Vol. 11006, 1100612, 2019
	
	\bibitem{zhao2021alike}
	X.~Zhao, X.~Wu, J.~Miao, W.~Chen, P.~C. Chen, and Z.~Li, Alike: Accurate and
	lightweight keypoint detection and descriptor extraction, arXiv
		preprint arXiv:2112.02906, 2021.
	
	\bibitem{zhao2021probabilistic}
	X.~Zhao, J.~Liu, X.~Wu, W.~Chen, F.~Guo, and Z.~Li, Probabilistic spatial
	distribution prior based attentional keypoints matching network, IEEE
		Trans. on Circuits and Systems for Video Technology, 2021.
	
	\bibitem{sparsetodense}
	F.~Ma, G.~V. Cavalheiro, and S.~Karaman, Self-supervised sparse-to-dense:
	{Self}-supervised depth completion from lidar and monocular camera, in
	{International} {Conference} on {Robotics} and
		{Automation}, 3288--3295, 2019
	
	\bibitem{van2019sparse}
	W.~Van~Gansbeke, D.~Neven, B.~De~Brabandere, and L.~Van~Gool, Sparse and
	noisy lidar completion with rgb guidance and uncertainty, in 16th international conference on machine vision applications, 1--6, 2019
	
	\bibitem{jaritz2018sparse}
	M.~Jaritz, R.~De~Charette, E.~Wirbel, X.~Perrotton, and F.~Nashashibi, Sparse
	and dense data with cnns: Depth completion and semantic segmentation, in
	International Conference on 3D Vision, 52--60, 2018
	
	\bibitem{tang2020learning}
	J.~Tang, F.-P. Tian, W.~Feng, J.~Li, and P.~Tan, Learning guided
	convolutional network for depth completion, IEEE Trans. on
		Image Processing, Vol.~30, 1116--1129, 2020.
	
	\bibitem{park2020non}
	J.~Park, K.~Joo, Z.~Hu, C.-K. Liu, and I.~So~Kweon, Non-local spatial
	propagation network for depth completion, in European Conference on Computer Vision, Glasgow, UK, 120--136, 2020
	
	\bibitem{penet}
	M.~Hu, S.~Wang, B.~Li, S.~Ning, L.~Fan, and X.~Gong, {PENet}: {Towards}
	{Precise} and {Efficient} {Image} {Guided} {Depth} {Completion}, in
	{IEEE International Conference on Robotics and Automation}, 2021.
	
	\bibitem{zhao2021adaptive}
	S.~Zhao, M.~Gong, H.~Fu, and D.~Tao, Adaptive context-aware multi-modal
	network for depth completion, IEEE Trans. on Image Processing,
	2021.
	
	\bibitem{wang20193d}
	T.-H. Wang, H.-N. Hu, C.~H. Lin, Y.-H. Tsai, W.-C. Chiu, and M.~Sun, 3d lidar
	and stereo fusion using stereo matching network with conditional cost volume
	normalization, in IEEE/RSJ International Conference on
		Intelligent Robots and Systems.x, 5895--5902, 2019
	
	\bibitem{kou2017intelligent}
	F.~Kou, Z.~Wei, W.~Chen, X.~Wu, C.~Wen, and Z.~Li, Intelligent detail
	enhancement for exposure fusion, IEEE Trans. on Multimedia,
	Vol.~20, No.~2, 484--495, 2018.
	
\end{thebibliography}
\end{document}